\documentclass{article}

\usepackage{times}
\usepackage{graphicx} 
\usepackage{subfigure}
\usepackage{amssymb,amsmath}
\usepackage{amssymb}
\usepackage{amsmath}
\usepackage{mathptmx}
\usepackage{mathrsfs}
\usepackage{amsfonts}
\usepackage{amsxtra}
\usepackage[T1]{fontenc}
\usepackage{multirow}
\usepackage{verbatim}
\usepackage{comment}
\usepackage{multicol}
\usepackage[final]{pdfpages}
\usepackage{bm}
\usepackage{natbib}

\usepackage{algorithm}
\usepackage{algorithmic}

\newcommand{\elle}{\mathit{l}}

\newcommand{\rx}{\mathtt{x}}

\newcommand{\ry}{\mathrm{y}}
\newcommand{\vx}{\mathbf{x}}
\newcommand{\vy}{\mathbf{y}}
\newcommand{\vz}{\mathbf{z}}
\newcommand{\Real}{I \! \! R}
\newcommand{\base}{\mathit{g}}

\newcommand{\sD}{\mathscr{D}}
\newcommand{\sY}{\mathscr{Y}}
\newcommand{\Realsp}{\Real_{+}^{*}}

\newcommand{\nmathbf}{\bm}

\def\bfgamma  {\nmathbf \gamma}
\def\bfdelta  {\nmathbf \delta}

\def\boldfacefake#1{\kern-4pt
    \hbox{ \mathsurround=0pt
    \hbox to 0.4pt{$#1$\hss}\hbox to 0.4pt{$#1$\hss}\hbox {$#1$}}}

\newcommand{\be}{\begin{eqnarray}}
\newcommand{\ee}{\end{eqnarray}}
\newcommand{\ba}{\begin{eqnarray*}}
\newcommand{\ea}{\end{eqnarray*}}

\newcommand{\reals}{\mbox{\rm I\kern-.20em R}}
\newcommand{\sreals}{\mbox{\small \rm I\kern-.20em R}}

\newtheorem{theorem0}{Theorem}
\newtheorem{lemma0}{Lemma}
\newtheorem{remark0}{Remark}
\newtheorem{fact0}{Fact}
\newtheorem{example0}{Example}
\newtheorem{definition0}{Definition}
\newtheorem{corollary0}{Corollary}
\newtheorem{proposition0}{Proposition}
\newtheorem{algorithmY}{Algorithm}
\newtheorem{conjecture0}{Conjecture}

\newenvironment{definition}{\begin{definition0}
\mbox{}}{\end{definition0}}

\makeatletter
\@tfor\next:=abcdefghijklmnopqrstuvwxyzABCDEFGHIJKLMNOPQRSTUVWXYZ\do{%
  \def\command@factory#1{%
    \expandafter\def\csname V#1\endcsname{\mathbb{#1}}
  }
 \expandafter\command@factory\next
}

\usepackage{hyperref}

\usepackage[accepted]{icml2015}

\icmltitlerunning{Submission and Formatting Instructions for ICML 2015}
\begin{document}

\twocolumn[
\icmltitle{Adaptive Random SubSpace Learning (RSSL) Algorithm for Prediction}

\icmlauthor{Mohamed Elshrif}{mme4362@rit.edu}
\icmladdress{Rochester Institute of Technology (RIT),
            102 Lomb Memorial Dr., Rochester, NY 14623 USA}
\icmlauthor{Ernest Fokou\'e}{epfeqa@rit.edu}
\icmladdress{Rochester Institute of Technology (RIT),
            102 Lomb Memorial Dr., Rochester, NY 14623 USA}

\icmlkeywords{boring formatting information, machine learning, ICML}

\vskip 0.3in
]

\begin{abstract}
We present a novel adaptive random subspace learning algorithm (RSSL) for prediction purpose. This new framework is flexible where it can be adapted with any learning technique. In this paper, we tested the algorithm for regression and classification problems. In addition, we provide a variety of weighting schemes to increase the robustness of the developed algorithm. These different wighting flavors were evaluated on simulated as well as on real-world data sets considering the cases where the ratio between features (attributes) and instances (samples) is large and vice versa. The framework of the new algorithm consists of many stages: first, calculate the weights of all features on the data set using the correlation coefficient and F-statistic statistical measurements. Second, randomly draw $n$ samples with replacement from the data set. Third, perform regular bootstrap sampling (bagging). Fourth, draw without replacement the indices of the chosen variables. The decision was taken based on the heuristic subspacing scheme. Fifth, call base learners and build the model. Sixth, use the model for prediction purpose on test set of the data. The results show the advancement of the adaptive RSSL algorithm in most of the cases compared with the synonym (conventional) machine learning algorithms.
\end{abstract}

\section{Introduction}
Given a dataset $\mathscr{D} = \{\vz_i=(\vx_i^\top,\ry_i)^\top,\,\,i=1,\cdots,n\}$, where
$\vx_i = (\rx_{i1},\cdots,\rx_{ip})^\top \in \mathscr{X}\subset \Real^p$ and  $\ry_i\in \mathscr{Y}$ are realizations of two random variables $X$ and $Y$ respectively, we seek to use the data $\mathscr{D}$ to build estimators $\widehat{f}$ of the underlying function $f$  for predicting the response $Y$ given the vector $X$ of explanatory variables. In keeping with the standard in statistical learning theory, we will measure the predictive performance of any given function $f$ using the theoretical risk functional given by
\begin{eqnarray}
\label{eq:risk:1}
\mathcal{R}(f) = \mathbb{E}[\ell(Y,f(X))] = \int_{\mathscr{X} \times \mathscr{Y}}{\ell(\vx,\ry) d P(\vx,\ry)},
\end{eqnarray}
with the ideal scenario corresponding to the universally best function defined by
\begin{eqnarray}
\label{eq:opt:f:1}
\widehat{f}^* =  {\tt arg} \,\underset{f}{\tt inf}\left\{\mathcal{R}(f)\right\}= {\tt arg} \,\underset{f}{\tt inf}\left\{\mathbb{E}[\ell(Y,f(X))]\right\}.
\end{eqnarray}
For classification tasks, the most commonly used loss function is the zero-one loss
$\ell(Y,f(X)) = 1_{\{Y \neq f(X)\}}$, for which the theoretical universal best defined in \eqref{eq:opt:f:1} is the Bayes classifier given
by $f^*(\vx) = {\tt arg}\underset{\ry \in \sY}{\tt max}\left\{\Pr[Y=\ry|\vx]\right\}$.
For regression tasks, the squared loss $\ell(Y,f(X)) = (Y-f(X))^2$ is by far the most commonly used, mainly because of the wide variety of statistical, mathematical and computational benefits it offers.  For regression under the squared loss, the universal best defined in \eqref{eq:opt:f:1} is also known theoretically known to be the conditional expectation of $Y$ given $X$, specifically given by
$f^*(\vx) = \mathbb{E}[Y|X=\vx]$. Unfortunately, these theoretically expressions of the best estimators cannot be realized in practice because the distribution function $P(\vx,\ry)$ of $(X,Y)$ defined on $\mathscr{X}\times\mathscr{Y}$ is unknown. To circumvent this learning challenge, one has to do essentially two foundational thing, namely: (a) choose a certain function class $\mathscr{F}$ (approximation) from which to search for the estimator $\widehat{f}$ of the true but unknown underlying $f$, (b) specify the empirical version of \eqref{eq:risk:1} based on the given sample $\mathscr{D}$, an use that empirical risk as the practical objective function. However, in this paper, we do not directly construct our estimating classification functions from the empirical risk. Instead, we build the estimators using other optimality criteria, and then compare their predictive performances using the average test error  $\mathtt{AVTE}(\cdot)$, namely
\begin{eqnarray}
    \label{eq:avte:1}
    \mathtt{AVTE}(\widehat{f}) =\frac{1}{R} \sum_{r=1}^{R} \left\{ \frac{1}{m} \sum_{t=1}^{m} \ell(\vy_{i_t}^{(r)}, \widehat{f}_{r}(\vx_{i_t}^{(r)}))\right\},
\end{eqnarray}
where  $\widehat{f}_{r}(\cdot)$ is the $r$-th  realization of the estimator $\widehat{f}(\cdot)$ built using the training portion of the split of $\mathscr{D}$ into training set and test set, and $\left(\vx_{i_t}^{(r)},\vy_{i_t}^{(r)}\right)$ is the $t$-th observation from the test set at the $r$-th random replication of the split of $\mathscr{D}$. In this paper, we consider both multiclass classification tasks with response space $\sY = \{1,2,\cdots, G\}$  and regression  tasks  with $\sY = \Real$, and we focus on learning machines from a function class
$\mathscr{F}$ whose members are ensemble learners in the sense of Definition \eqref{def:ensemble:1}. In machine learning, in order to improve the accuracy of a regression function, or a classification function, scholars tend to combine multiple estimators because it has been proven both theoretically and empirically \cite{Tumer95classifiercombining, Tumer99decimatedinput} that an appropriate
combination of good base learners leads to a reduction in prediction error. This technique is known as ensemble learning (aggregation). In spite of the underlying algorithm used, the ensemble learning technique  most of the time (on average) outperforms the single learning technique, especially for prediction purposes \cite{vanWezel2007436}. There are many approaches of performing ensemble learning. Among these, there are two popular ensemble learning techniques, bagging \cite{Breiman_1996} and boosting \cite{Freund1995256}. Many variants of these two techniques have been studied previously such as random forest \cite{Breiman:2001:RF:570181.570182} and AdaBoost \cite{Freund1997119} and applied in a prediction problem. Our proposed method
belongs to the subclass of ensemble learning methods known as random subspace learning.
\begin{definition}
\label{def:ensemble:1}
Given an ensemble $\mathscr{H} = \{h_1, h_2, \cdots , h_L \}$ of base learners $h_\elle:\mathscr{X}\longrightarrow \mathscr{Y}$,  with
relative weight $\alpha_{\elle} \in \Realsp$ (usually $\alpha_\elle \in (0,1)$ for convex aggregation),
the ensemble representation $f^{(L)}$ of the underlying function $f$ is given by the aggregation (weighted sum)
\begin{eqnarray}
\label{eq:ensemble:1}
{f}^{(L)}(\cdot) = \sum_{\elle=1}^L{{\alpha}_{\elle}{h}_{\elle}(\cdot)}.
\end{eqnarray}
\end{definition}

A question naturally arises as to how the ensemble $\mathscr{H}$ is chosen, and how the weights are determined.
Bootstrap Aggregating also known as bagging \cite{Breiman_1996}, boosting \cite{Freund96experimentswith}, random forests \cite{Breiman:2001:RF:570181.570182}, and bagging with subspaces \cite{Panov_2007} are all predictive learning methods
based on the ensemble learning principle for which the ensemble is built from the provided data set $\mathscr{D}$ and the weights are typically taken to be equal. In this paper, we focus on learning tasks involving high dimension low sample size (HDLSS) data, and we  further zero-in on those data sets for which the number of explanatory variables $p$ is substantially larger than the sample size $n$. As our main contribution in this paper, we introduce, develop and apply a new adaptation of the theme of random subspace learning \cite{Ho:1998:RSM:284980.284986} using the traditional multiple linear regression (MLR) model as our base learner in regression and the generalized linear model (GLM) as a base learner in classification.
Some applications by nature posses few instances (small $n$) with large number of features ($p \ggg n$) such as fMRI \cite{5405643} and DNA microarrays \cite{Bertoni2005535} data sets. It is hard for a traditional (conventional) algorithm to build a regression model, or to classify the data set when it possesses a very small instances to features ratio. The prediction problem becomes even more difficult when this huge number of features correlated are highly correlated, or irrelevant for the task of building such a model, as we will show later in this paper. Therefore, we harness the power of our proposed adaptive subspace learning technique to guide the choice/selection of good candidate features from the data set, and therefore select the best base learners, and ultimately the ensemble yielding the lowest possible prediction error. In most typical random subspace learning algorithms, the features are selected according to an {\it equally likely} scheme. The question then arises as to whether one can {\it devise a better scheme to choose the candidate features for efficiently with some predictive benefits}. On the other hand, it is interesting to assess  {\it the accuracy of our proposed algorithm under different levels of the correlation of the features}. The answer to this question constitutes one of the central aspect of our proposed method, in the sense {\it we explore a variety of weighting scheme for choosing the features, most of them (the schemes) based on statistical measures of relationship between the response variable and each explanatory variable}. As the computational section will reveal, the weighting schemes proposed here lead to a substantially improvement in predictive performance of our method over random forest on all but one data set,
arguably due to the fact that our method because it leverages the accuracy of the learning algorithm through selecting many good models ({\it since
the weighting scheme allows good variables to be selected more often and therefore leads to near optimal base learners}).


\section{Related Work}
Traditionally, in a prediction problem, a single model is built based on the training set and the prediction is decided based solely on this single fitted model. However, in bagging, bootstrap samples are taken from the data set, then, for each instance, the model is fitted. Finally, the prediction is made based on the average of all bagged models. Mathematically, the prediction accuracy for the constructed model using bagging outperforms the traditional model and in the worst case it has the same performance. However, it must be said that it depends on the stability of the modeling procedure. It turns out that bagging reduces the variance without affecting the bias, thereby leading to an overall reduction in prediction error, and hence its great appeal.
Any set of predictive models can be used as an ensemble in the sense defined earlier. There are many ensemble learning approaches. These approaches could be categorized into four classes: (1) algorithms that use heterogeneous predictive models such as stacking \cite{Wolpert92stackedgeneralization}. (2) algorithms that manipulate the instances of the data sets such as bagging \cite{Breiman_1996}, boosting \cite{Freund96experimentswith}, random forests \cite{Breiman:2001:RF:570181.570182}, and bagging with subspaces \cite{Panov_2007}. (3) algorithms that maniplulate the features of the data sets such as random forests \cite{Breiman:2001:RF:570181.570182}, random subspaces \cite{Ho:1998:RSM:284980.284986}, and bagging with subspaces \cite{Panov_2007}. (4) algorithms that manipulate the learning algorithm such as random forests \cite{Breiman:2001:RF:570181.570182}, neural networks ensemble \cite{Hansen:1990:NNE:628297.628429}, and extra-trees ensemble \cite{Geurts_2006}.
Since our proposed  algorithm manipulates both the instances and features of the data sets,  we will focus on the algorithms in the second and third categories \cite{Breiman_1996, Breiman:2001:RF:570181.570182, Panov_2007, Ho:1998:RSM:284980.284986}.

Bagging \cite{Breiman_1996}, or bootstrap aggregating is an ensemble learning method that generates multiple predictive models. These models are based on performing bootstrap replicates of the learning (training) data set and utilizing from each replicate to build a separate predictive model. The bootstrap sample is attained through randomly (uniformly) sampling with replacement from instances of the training data set. The decision is made based on averaging the predictor classifiers in regression task and taking the majority vote in classification task.  Bagging tend to decrease the variance and keeps the bias as in the case of a single classifier. The bagging accuracy increases when the applied learner is unstable, which means that for any small fluctuation on the training data set causes large impact on the test data set  such as trees \cite{Breiman_1996}.
Random forests \cite{Breiman:2001:RF:570181.570182}, is an ensemble learning method that averages the prediction results from multiple independent predictor (tree) models. It also performs bootstrap replicates, like bagging \cite{Breiman_1996}, to construct different predictors. For each node of the tree, randomly selecting subset of the attributes. It is considered to improve over bagging through {\it de-correlating the trees. Choose the best attribute from the selected subset}. As \cite{Denil2014a} mentions that when building a random tree, there are three issues that should be decided in advance; (1) the leafs splitting method, (2) the type of predictor, and 3- the randomness method.
Random subspace learning \cite{Ho:1998:RSM:284980.284986}, is an ensemble learning method that constructs base models based on different features. It chooses a subset of features and then learns the base model depending only on these features. The random subspaces reaches the highest accuracy when the number of features is large as well as the number of instances. In addition, it performs good when there are redundant features on the data set.
Bagging subspaces \cite{Panov_2007}, is an ensemble learning method that combines both the bagging \cite{Breiman_1996} and random subspaces \cite{Ho:1998:RSM:284980.284986} learning methods. It generates a bootstrap replicates of the training data set, in the same way as bagging. Then, it randomly chooses a subset from the features, in the same manner as random subspaces. It outperforms the bagging and random subspaces. Also, it is found to yield the same performance as random forests in case of using decision tree as a base learner.
In the simulation part of this paper, we aim to answer the following research questions: (1) Is the performance of the adaptive random subspace learning (RSSL) better than the performance of single classifiers? (2) What is the performance of the adaptive RSSL compared to the most widely used classifier ensembles? (3) Is there a theoretical explanation as to why adaptive RSSL works well for most of the simulated and real-life data sets? (4) How does  adaptive RSSL perform on different parameter settings and with various percentages of the instance-to-feature ratio (IFR)? (5) How does the correlation between features affect the predictive performance of adaptive RSSL?

\section{Adaptive RSSL}
In this section, we present an adaptive random subspace learning algorithm for the prediction problem. We start with the formulation of the problem, followed by our suggested solution (proposed algorithm) to tackle (handle) it. A crucial step of assessing the candidate features for building the models is explained in detail. Finally, we elucidate the strength of the new algorithm, from a theoretical perspective.

\subsection{Problem formulation}
As we said earlier our proposed method belongs to the category of random subspace learning where
each base learner is constructed using a bootstrap sample and a subset of the original $p$ features.
The main difference here is that we use base learners that are typically considered not to lead to
any improvement when aggregated, and we also select features using weighting schemes that inspired
for the strength of the relationship between each feature and the response (target).
Each base learner is driven by the subset $\{j_1^{(\elle)},\cdots,j_d^{(\elle)}\} \subset \{1,2,\cdots,p\}$ of $d$ variables of predictors that are randomly select to build it, and the subsample $\sD^{(\elle)}$ drawn with replacement from $\sD$. For notational convenience, we use vectors of indicator variables to denote these two important quantities. The {\it sample indicator} $\bfdelta^{(\elle)} = (\delta_1^{(\elle)},\cdots, \delta_n^{(\elle)}) \in \{0,1\}^n$, where
$$
\delta_i^{(\elle)} = \left\{\begin{array}{ll} 1 & \texttt{if} \,\, \vz_i \in \sD^{(\elle)}\\
0 & \texttt{otherwise}\end{array}\right.
$$
The {\it variable indicator}  $\bfgamma^{(\elle)} = (\gamma_1^{(\elle)},\cdots, \gamma_p^{(\elle)}) \in \{0,1\}^p$, where
$$
\gamma_j^{(\elle)} = \left\{\begin{array}{ll} 1 & \texttt{if} \,\, j \in \{j_1^{(\elle)},\cdots,j_d^{(\elle)}\}\\
0 & \texttt{otherwise}\end{array}\right.
$$
The estimator of the $\elle$th base learner can therefore fully and unambiguously denoted by $\widehat{\base}^{(\elle)}(\cdot; \bfgamma^{(\elle)}, \bfdelta^{(\elle)})$ which we refer to as $\widehat{\base}^{(\elle)}(\cdot)$ for notational simplicity. Each $\delta_j^{(\ell)}$ is chosen according to one of the weighting schemes. To increase the strength of the developed algorithm, we introduce a weighting scheme procedure to select the important features, which facilitates building a proper model and leverage the prediction accuracy. Our weighting schemes are
\begin{itemize}
\item{Correlation coefficient:} We measure the strength of the association between each feature vector and the response (target), and take the square of the resulting sample correlation
\item{F-statistic:} For classification tasks especially, we use the observed F-statistic resulting from the analysis of variance with $\rx_j$ as the response and the class label the treatment group.
\end{itemize}
Using the ensemble $\Big\{\widehat{\base}^{(\elle)}(\cdot),\, \elle=1,\cdots, L\Big\}$ we form the ensemble estimator of class Membership as
\begin{eqnarray*}
\widehat{f}^{(L)}(\vx) =  {\tt arg} \,\underset{\ry \in \mathscr{Y}}{\tt max}\left\{\sum_{\elle=1}^{L}{\left({\bf 1}_{\{\ry=\widehat{\base}^{(\elle)}(\vx)\}}\right)}\right\},
\end{eqnarray*}
and the ensemble estimator of regression response as
$$
 \widehat{{f}}^{(L)}(\vx^*) = \frac{1}{L}\sum_{\elle=1}^{L}{\hat{\base}^{(\elle)}(\vx^*)}.
$$

\section{Experiments}

We used a collection of simulated and real-world data sets for our experiments. In addition, we used real-world data sets from previous papers, which aim to solve the same problem, for comparison purpose. We report the mean square error (MSE) for each individual algorithm and task purposes, i.e., regression, or classification.

\subsection{Simulated data sets}

We designed our artificial data sets to fit six scenarios based on the factors, which are the dimensionality of the data (number of features), the number of sample size ((number of instances), and the correlation of the data.

\subsection{Real data sets}

We benefit from the public repository of the UCI University real-life data sets in our paper. For the purposes of consistency and completeness, we choose the real data sets that carries different characteristics in terms of the number of instances and the number of features along with variety of applications. The real data sets can be represented based on the task as follows:

\begin{figure}[ht]
\begin{center}
\centerline{\includegraphics[width=\columnwidth]{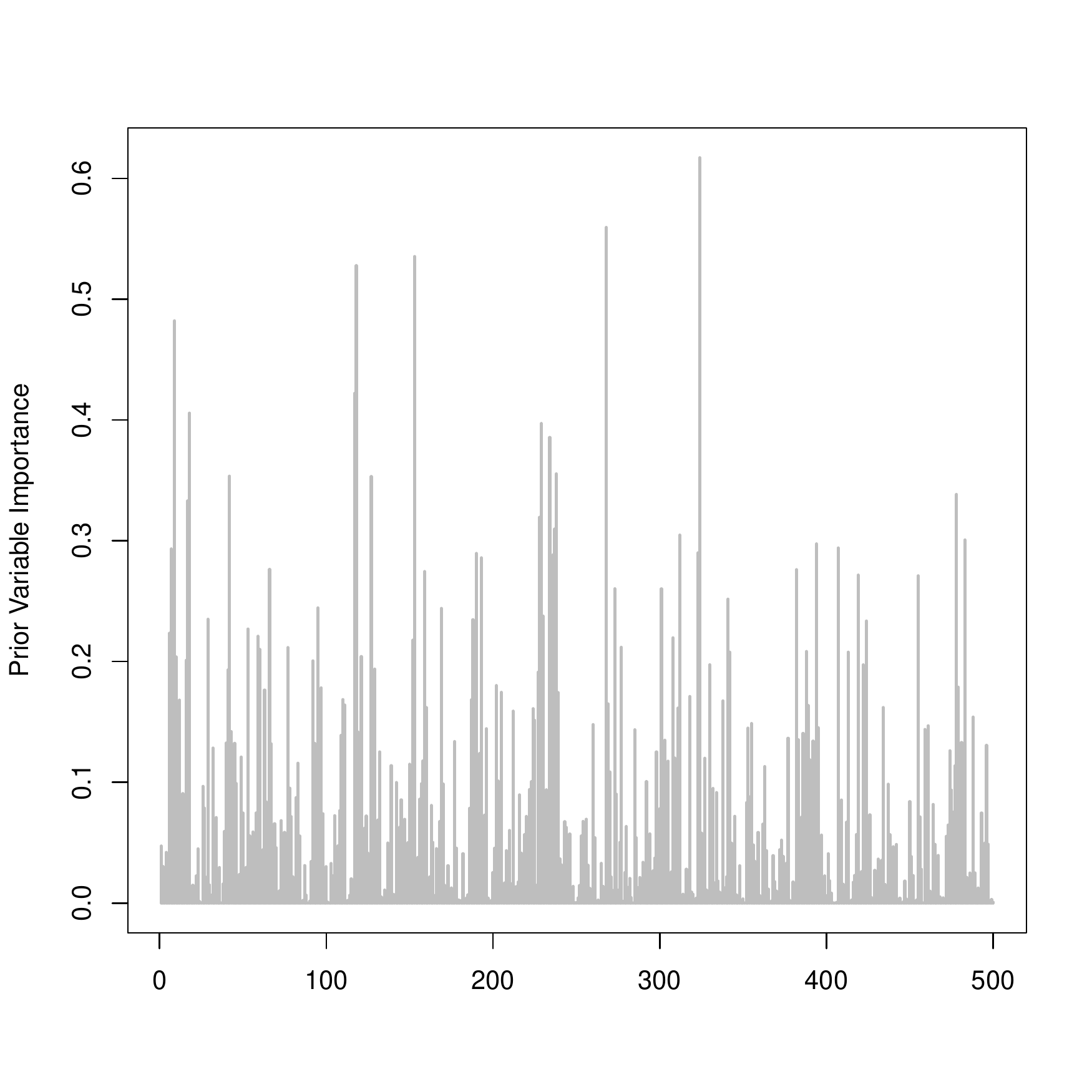}}
\caption{Prior Feature Importance: A representative simulation results for regression analysis on synthetic dataset of scenario with number of instances n=25, number of features p=500, correlation coefficient $\rho$=0.5, number of learners=450, and number of replications=100.}
\label{Figure 1}
\end{center}
\end{figure}

\begin{figure}[ht]
\begin{center}
\centerline{\includegraphics[width=\columnwidth]{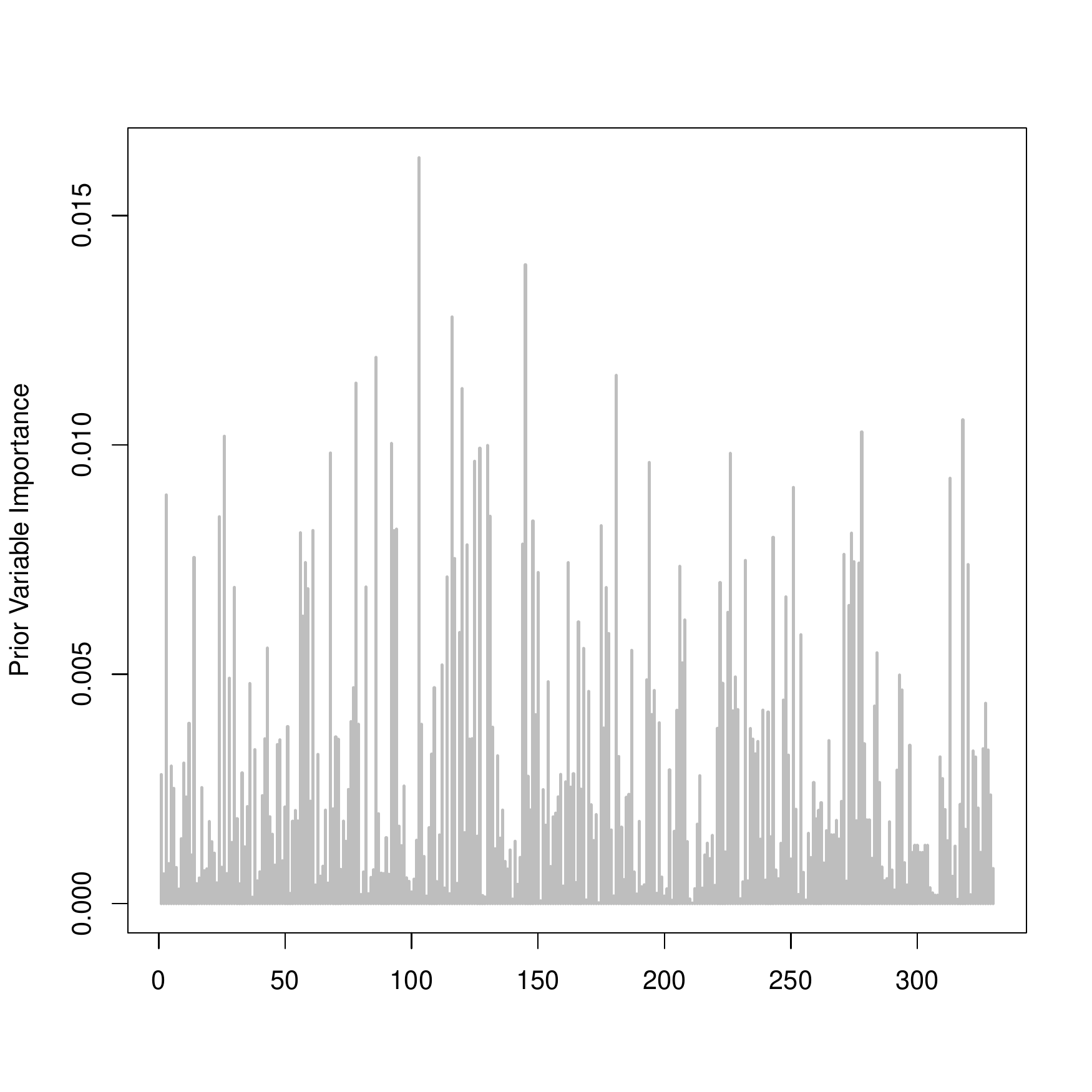}}
\caption{Prior Feature Importance: A representative simulation results for classification analysis on real dataset of Lymphoma disease.}
\label{Figure 2}
\end{center}
\end{figure}

\begin{figure}[ht]
\vskip 0.2in
\begin{center}
\centerline{\includegraphics[width=\columnwidth]{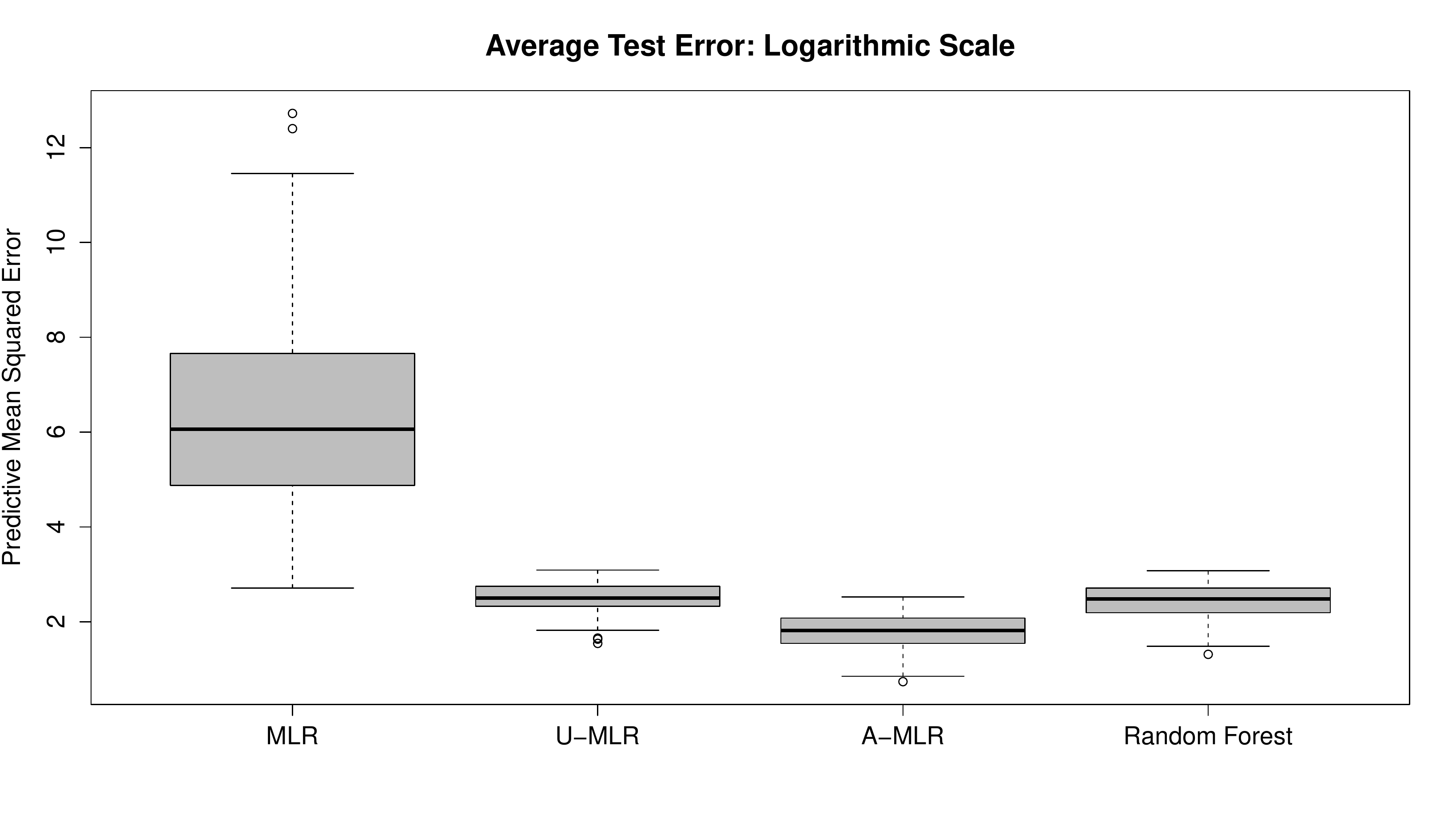}}
\caption{A representative results of synthetic dataset of scenario with number of instances n=50, number of features p=1000, correlation coefficient $\rho$=0.05, number of learners=450, and number of replications=100. We used the correlation weighting scheme for regression analysis on logarithmic scale.}
\label{Figure 3}
\end{center}
\vskip -0.2in
\end{figure}

\begin{figure}[ht]
\vskip 0.2in
\begin{center}
\centerline{\includegraphics[width=\columnwidth]{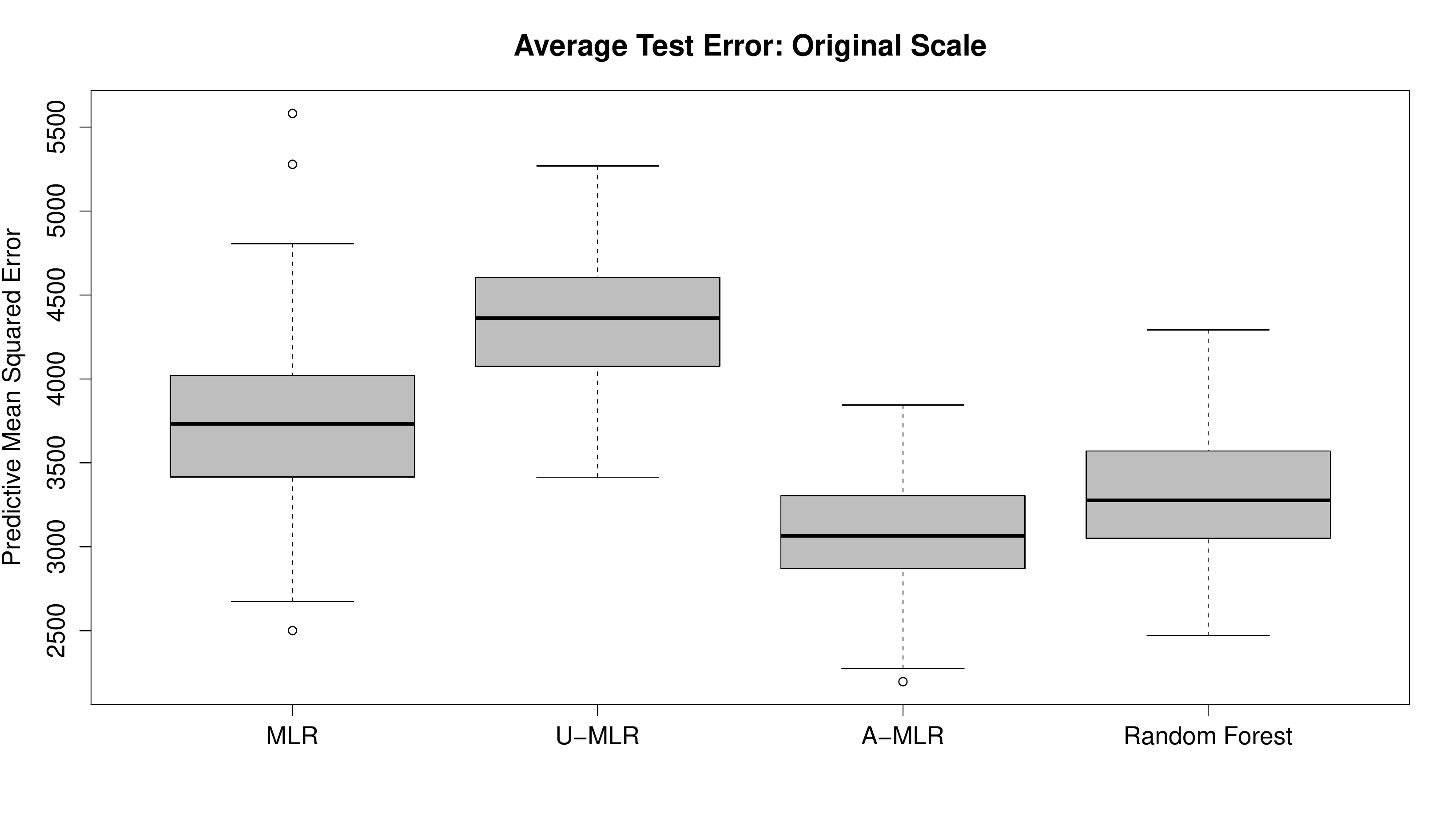}}
\caption{A representative results of Diabetes interaction real dataset with correlation weighting scheme for regression analysis on original scale.}
\label{Figure 4}
\end{center}
\vskip -0.2in
\end{figure}

\begin{figure}[ht]
\vskip 0.2in
\begin{center}
\centerline{\includegraphics[width=\columnwidth]{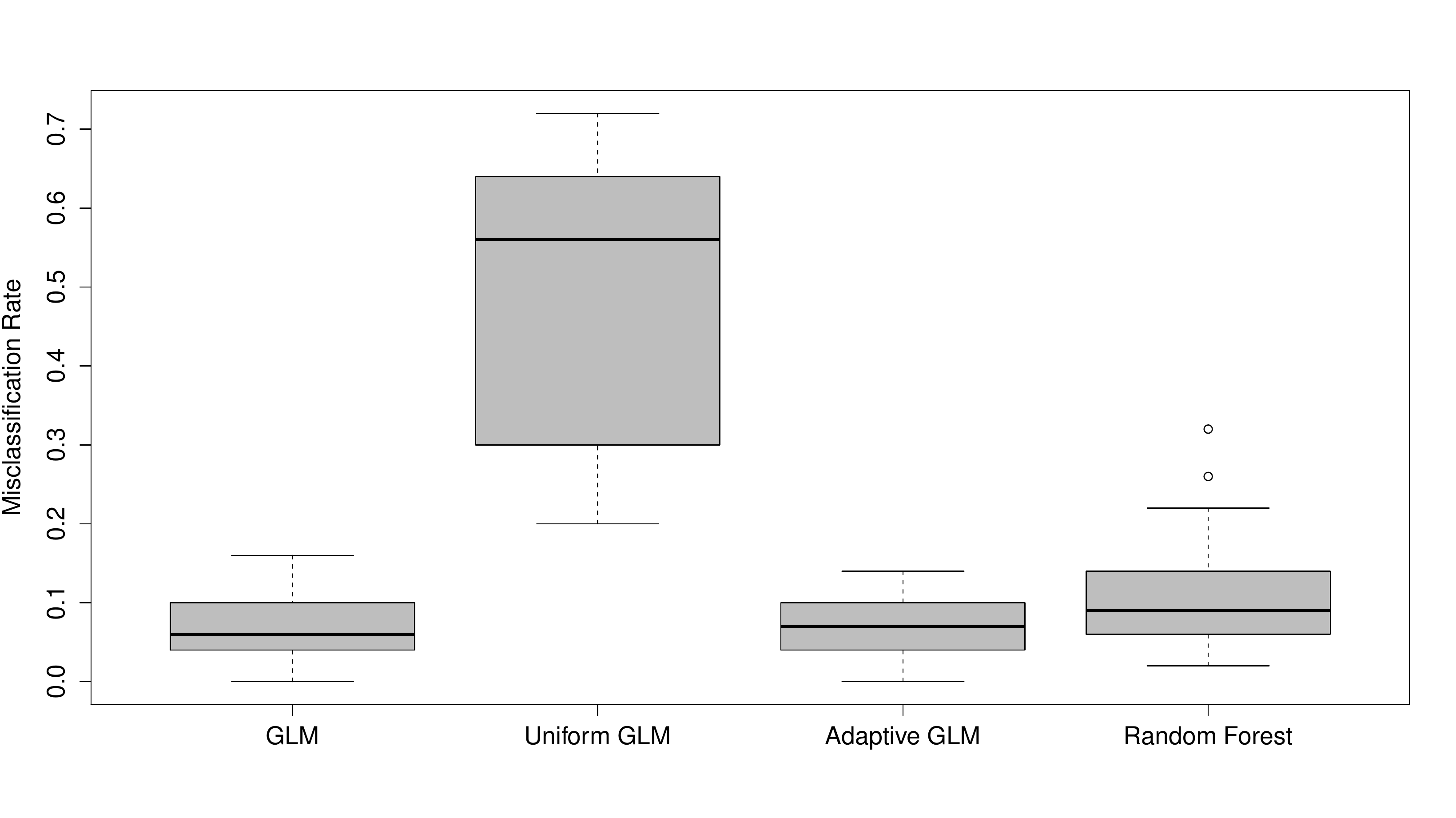}}
\caption{A representative results on synthetic dataset of scenario with number of instances n=200, number of features p=25, correlation coefficient $\rho$=0.05, number of learners=450, and number of replications=100. We used F-statistics weighting scheme for classification analysis.}
\label{Figure 5}
\end{center}
\vskip -0.2in
\end{figure}

\begin{figure}[ht]
\vskip 0.2in
\begin{center}
\centerline{\includegraphics[width=\columnwidth]{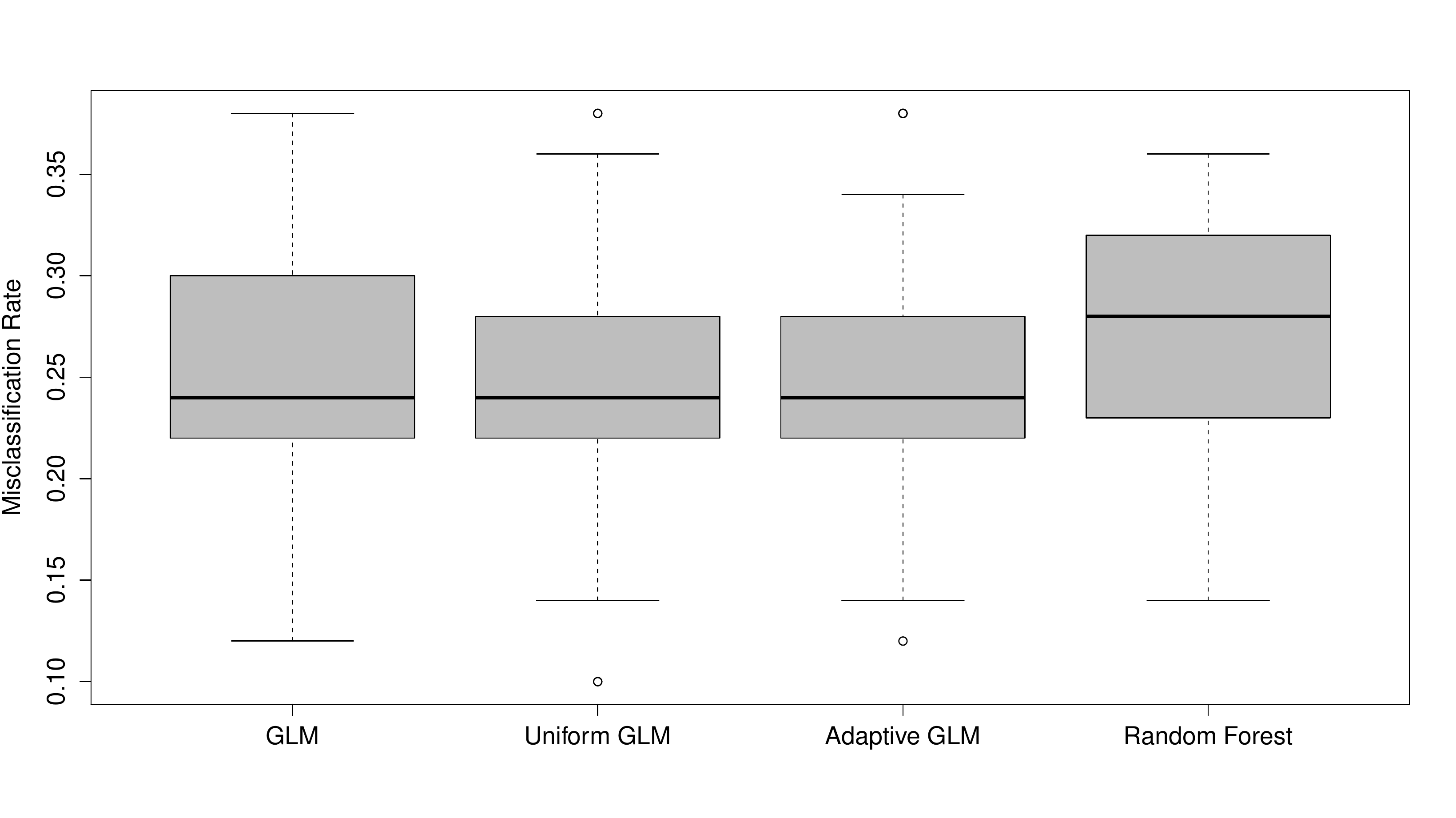}}
\caption{A representative results of the Diabetes in Pima Indian Women real dataset with F-statistics weighting scheme for classification analysis.}
\label{Figure 6}
\end{center}
\vskip -0.2in
\end{figure}


\begin{table*}[t]
\caption{Regression Analysis: Mean Square Error (MSE) for different machine learning algorithms on various scenarios of synthetic data sets.}
\label{sample-table}
\vskip 0.15in
\begin{center}
\begin{small}
\begin{sc}
\begin{tabular}{lcccccccc}
\hline
\abovespace\belowspace
Weighting & n & p & $\rho$ & MLR & Uniform MLR & Adaptive MLR & RF & Better? \\
\hline
\abovespace
\multirow{6}{*}{Correlation}          & 200 & 25 & 0.05 & 5.69$\pm$0.89 & 14.50$\pm$2.63 & 4.60$\pm$0.706  & 9.81$\pm$1.86 & $\surd$ \\
                                                    & 200 & 25 & 0.5 & 4.78$\pm$0.81 & 11.67$\pm$2.55 & 4.77$\pm$0.94 & 8.46$\pm$1.97 & $\surd$ \\
                                                    & 25 & 200 & 0.05 & 974.37$\pm$5.e3 & 18.35$\pm$6.92 & 8.10$\pm$3.86 & 18.56$\pm$7.24 & $\surd$ \\
                                                    & 25 & 200 &  0.5 & 5.e3$\pm$5.e4 & 18.83$\pm$8.72 & 8.27$\pm$5.24 & 18.18$\pm$8.65 & $\surd$ \\
                                                    & 50 & 1000 & 0.05 & 2.e4$\pm$1.e5 & 28.36$\pm$11.51 & 12.38$\pm$5.91 & 27.92$\pm$11.78 & $\surd$ \\
                                                    & 1000 & 50 & 0.05  & 4.66$\pm$0.34  & 16.62$\pm$1.37 & 4.33$\pm$0.33 & 6.73$\pm$0.62 & $\surd$ \\
\hline
\abovespace
\multirow{6}{*}{F-statistics}        & 200 & 25 & 0.05 & 5.04$\pm$0.79 & 14.42$\pm$2.67 & 4.48$\pm$0.74 & 8.75$\pm$1.76 & $\surd$ \\
                                                  & 200 & 25 & 0.5 & 4.49$\pm$0.76 & 12.06$\pm$2.04 & 5.51$\pm$1.09 & 8.33$\pm$1.59 & $\surd$ \\
                                                  & 25 & 200 & 0.05 & 3.e4$\pm$2.e5 & 17.77$\pm$9.15 & 5.81$\pm$4.10 & 15.81$\pm$8.55 & $\surd$ \\
                                                  & 25 & 200 &  0.5 & 1.e4$\pm$1.e5 & 23.09$\pm$16.06 & 12.53$\pm$10.27 & 24.11$\pm$16.31 & $\surd$ \\
                                                  & 50 & 1000 & 0.05 & 4.e5$\pm$3.e6 & 16.65$\pm$5.38 & 7.65$\pm$2.83 & 15.54$\pm$5.31 & $\surd$ \\
                                                  & 1000 & 50 & 0.05 & 4.19$\pm$0.33 & 15.97$\pm$1.15 & 3.90$\pm$0.30 & 6.24$\pm$0.55 & $\surd$ \\
\hline
\hline
\end{tabular}
\end{sc}
\end{small}
\end{center}
\vskip -0.1in
\end{table*}

\begin{table*}[t]
\caption{Regression Analysis: Mean Square Error (MSE) for different machine learning algorithms on real data sets.}
\label{sample-table}
\vskip 0.15in
\begin{center}
\begin{small}
\begin{sc}
\begin{tabular}{lcccccc}
\hline
\abovespace\belowspace
Data Set & Weighting & MLR & Uni. MLR & Adap. MLR & RF & Better? \\
\hline
\abovespace
\multirow{2}{*}{BodyFat} & correlation & 17.41$\pm$2.69 & 23.59$\pm$3.71 & 19.25$\pm$3.06 & 19.72$\pm$3.18 & $\times$ \\
                                        & F-statistics & 17.06$\pm$2.50 & 23.07$\pm$3.46 & 17.46$\pm$2.65 & 19.51$\pm$2.99 & $\times$ \\
\hline
\abovespace
\multirow{2}{*}{Attitude} & correlation & 74.12$\pm$32.06 & 80.35$\pm$34.40 & 58.49$\pm$20.21 & 88.72$\pm$35.97 & $\surd$ \\
                                      & F-statistics & 75.19$\pm$36.63 & 74.71$\pm$33.17 & 51.84$\pm$15.19 & 82.21$\pm$35.58 & $\surd$ \\
\hline
\abovespace
\multirow{2}{*}{Cement} & correlation & 10.76$\pm$7.25 & NA & 19.92$\pm$15.98 & 75.91$\pm$56.05 & $\times$ \\
                                      & F-statistics & 11.07$\pm$8.55 & NA & 24.27$\pm$18.27 & 62.20$\pm$46.53 & $\times$ \\
\hline
\abovespace
\multirow{2}{*}{Diabetes 1} & correlation & 2998.13$\pm$322.37 & 3522.30$\pm$311.81 & 3165.74$\pm$300.86 & 3203.94$\pm$311.94 & $\times$ \\
                                      & F-statistics & 2988.32$\pm$341.20 & 3533.45$\pm$375.38 & 3133.60$\pm$324.75 & 3214.11$\pm$318.6931 & $\times$ \\
\hline
\abovespace
\multirow{2}{*}{Diabetes 2} & correlation & 3916.98$\pm$782.35 & 4244.00$\pm$390.29 & 3016.54$\pm$285.89 & 3266.50$\pm$324.82 & $\surd$ \\
                                      & F-statistics & 3889.00$\pm$679.55 & 4306.76$\pm$419.66 & 3076.77$\pm$338.08 & 3326.28$\pm$382.37 & $\surd$ \\
\hline
\abovespace
\multirow{2}{*}{Longley} & correlation & 0.21$\pm$0.13 & 0.62$\pm$0.36 & 0.49$\pm$0.29 & 1.54$\pm$0.92 & $\times$ \\
                                      & F-statistics & 0.22$\pm$0.13 & 0.66$\pm$0.42 & 0.49$\pm$0.29 & 1.63$\pm$1.04 & $\times$ \\
\hline
\abovespace
\multirow{2}{*}{Prestige} & correlation & 66.68$\pm$15.31 & 73.32$\pm$14.77 & 64.87$\pm$13.93 & 55.96$\pm$11.83 & $\times$ \\
                                      & F-statistics & 65.77$\pm$15.96 & 72.33$\pm$16.64 & 63.27$\pm$14.71 & 56.02$\pm$12.66 & $\times$ \\

\hline
\hline
\end{tabular}
\end{sc}
\end{small}
\end{center}
\vskip -0.1in
\end{table*}

\begin{table*}[t]
\caption{Classification Analysis: MisClassification Rate (MCR) for different machine learning algorithms on various scenarios of simulated data sets.}
\label{sample-table}
\vskip 0.15in
\begin{center}
\begin{small}
\begin{sc}
\begin{tabular}{lccccccccr}
\hline
\abovespace\belowspace
Weighting  & n & p & $\rho$ & GLM & Uni. GLM & Adap. GLM & RF & Better? \\
\hline
\abovespace
\multirow{5}{*}{F-statistics} & 200 & 25 & 0.05 & 0.070$\pm$0.033 & 0.486$\pm$0.172 & 0.071$\pm$0.032 & 0.101$\pm$0.053 & $\surd$ \\
                                           & 200 & 25 & 0.5 & 0.140$\pm$0.045 & 0.498$\pm$0.221 & 0.138$\pm$0.043 & 0.136$\pm$0.058 & $\times$ \\
                                           & 50 & 200 & 0.05 & 0.102$\pm$0.093 & 0.673$\pm$0.123 & 0.100$\pm$0.092 & 0.320$\pm$0.103 & $\surd$ \\
                                           & 50 & 200 &  0.5 & 0.058$\pm$0.141 & 0.346$\pm$0.346 & 0.049$\pm$0.121 & 0.178$\pm$0.188 & $\surd$ \\
                                           & 50 & 1000 & 0.05 & 0.033$\pm$0.064 & 0.522$\pm$0.158 & 0.034$\pm$0.062 & 0.409$\pm$0.114 & $\surd$ \\
                                           & 1000 & 50 & 0.05 & 0.130$\pm$0.019 & 0.643$\pm$0.028 & 0.130$\pm$0.019 & 0.167$\pm$0.024 & $\surd$ \\
\hline
\hline
\end{tabular}
\end{sc}
\end{small}
\end{center}
\vskip -0.1in
\end{table*}

\begin{table*}[t]
\caption{Classification Analysis: MisClassification Rate (MCR) for different machine learning algorithms on real data sets.}
\label{sample-table}
\vskip 0.15in
\begin{center}
\begin{small}
\begin{sc}
\begin{tabular}{lcccccccr}
\hline
\abovespace\belowspace
Data Set & W. S.  & GLM & Uni. GLM & Adap. GLM & RF & Better? \\
\hline
\abovespace
Diabetes in Pima & F-stat & 0.274$\pm$0.071 & 0.249$\pm$0.051 & 0.255$\pm$0.051 & 0.269$\pm$0.050 & $\surd$ \\
\hline
Prostate Cancer & F-stat & 0.425$\pm$0.113 & 0.355$\pm$0.093 & 0.332$\pm$0.094 & 0.343$\pm$0.098 & $\surd$ \\
\hline
Golub Leukemia & F-stat & 0.427$\pm$ & 0.023$\pm$ & 0.021$\pm$ & 0.023$\pm$ & $\surd$ \\
\hline
Diabetes & F-stat & 0.034$\pm$0.031 & 0.068$\pm$0.039 & 0.038$\pm$0.034 & 0.031$\pm$0.029 & $\times$ \\
\hline
Lymphoma & F-stat & 0.248$\pm$0.065 & 0.057$\pm$0.034 & 0.046$\pm$0.029 & 0.082$\pm$0.046 & $\surd$ \\
\hline
Lung Cancer & F-stat & 0.113$\pm$0.051 & 0.038$\pm$0.023 & 0.037$\pm$0.024 & 0.051$\pm$0.030 & $\surd$ \\
\hline
Colon Cancer & F-stat & 0.296$\pm$0.124 & 0.168$\pm$0.095 & 0.124$\pm$0.074 & 0.199$\pm$0.106 & $\surd$ \\
\hline
\hline
\end{tabular}
\end{sc}
\end{small}
\end{center}
\vskip -0.1in
\end{table*}


\begin{table}[t]
\caption{Summary of the regression and classification real data sets.}
\label{sample-table}
\vskip 0.15in
\begin{center}
\begin{small}
\begin{sc}
\begin{tabular}{lccc}
\hline
\abovespace\belowspace
Data set & $\#$ inst. & $\#$ feat. & IFR ratio \\
\hline
regression &  &  &  \\
\hline
Bodyfat & 252 & 14 & 1,800\% \\
\hline
attitude & 30 & 7 & 428.5\% \\
\hline
Cement & 13 & 5 & 260\% \\
\hline
Diabetes 1 & 442 & 11 & 4,018\% \\
\hline
Diabetes 2 & 442 & 65 & 680\% \\
\hline
Longley & 16 & 7 & 228.5\% \\
\hline
Prestige & 102 & 5 & 2,100\% \\
\hline
classification &  &  &  \\
\hline
Diabetes in Pima & 200 & 8 & 2,500\% \\
\hline
Prostate cancer & 79 & 501 & 15.8\% \\
\hline
Leukemia & 72  & 3572  & 2.0\% \\
\hline
Diabetes & 145 & 4 & 3,625\% \\
\hline
Lymphoma & 180 & 662 & 27.2\%  \\
\hline
\hline
\end{tabular}
\end{sc}
\end{small}
\end{center}
\vskip -0.1in
\end{table}

\begin{figure}[ht]
\vskip 0.2in
\begin{center}
\includegraphics[trim = 93mm 30mm 75mm 32mm, clip, width=\columnwidth]{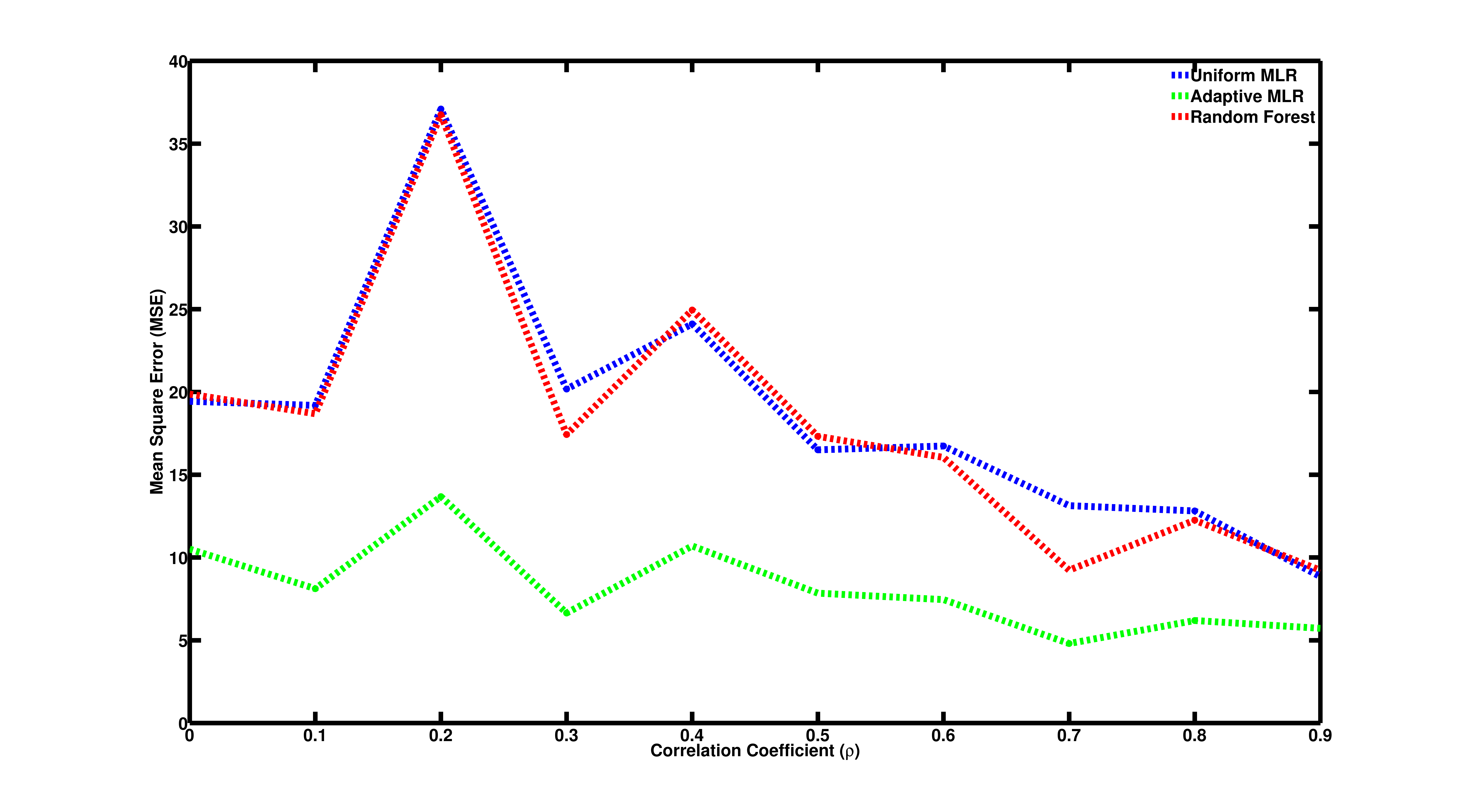}
\caption{A representative results that exhibits the relationship between mean square error (MSE) and correlation coefficient ($\rho$) for different algorithms  on synthetic dataset with correlation weighting scheme for regression analysis when p$\gg$n.}
\label{Correlation-regression-results}
\end{center}
\vskip -0.2in
\end{figure}

\begin{figure}[ht]
\vskip 0.2in
\begin{center}
\includegraphics[trim = 93mm 30mm 75mm 32mm, clip, width=\columnwidth]{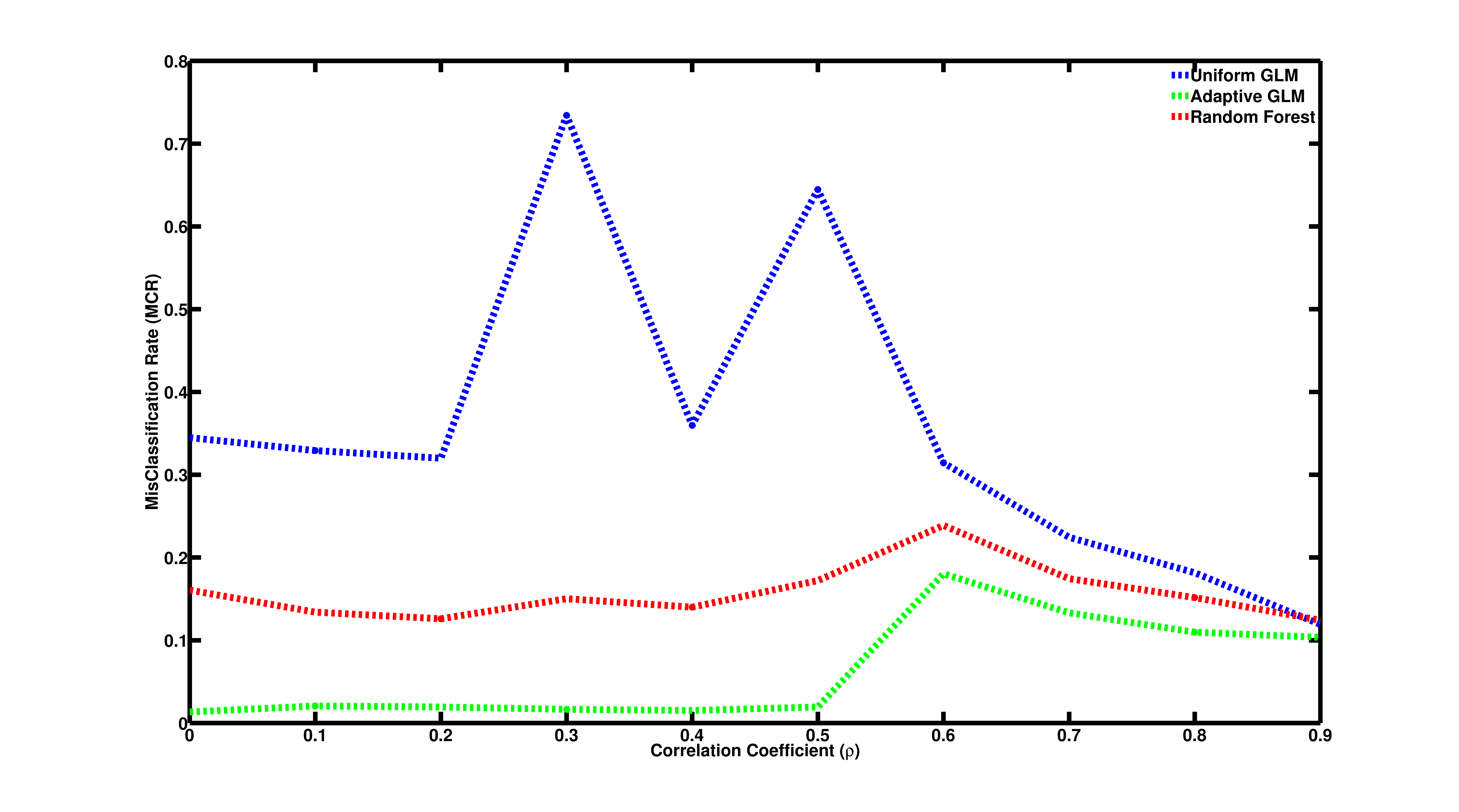}
\caption{A representative results that exhibits the relationship between mean square error (MSE) and correlation coefficient ($\rho$) for different algorithms  on synthetic dataset with F-statistics weighting scheme for classification analysis when n$\gg$p.}
\label{Correlation-classification-results}
\end{center}
\vskip -0.2in
\end{figure}


To elucidate the performance of our developed model, we compare the accuracy of the RSSL with random forest and ... on the same real data sets they used before.

\section{Discussion}

As revealed (experienced) from our experiments on synthetic data sets that when the number of selected features is higher than 15-20 (for our particular dataset) yields ensemble classifiers that are highly accurate and stable. The reason for this is that only if the number of voters is Òlarge enoughÓ does the random process of attribute selection yield suKcient number of qualitatively different classifiers that ensure high accuracy and stability of the ensemble.

how many bootstrap replications are useful?
The evidence both experimental and theoretical is that bagging can push a good but unstable procedure a significant step towards optimality.
why the training set in real dataset was chosen to be large and in simulated dataset the test set used to be large?
The bootstrap sample was repeated 50 times.
The random division of the data is repeated 100 times.
Choosing between these two strategies is not an easy task since it involves a trade-off between bias and estimation variance over the forecast horizon.

Even though that our developed adaptive RSSL algorithm outperforms many classifier ensembles. It has limitations where this new algorithm can not deal with data set that has categorical features. Instead it necessities to encode these features numerically. Also, the algorithm is not  designed to classify data sets with multiple classes. Moreover, the adaptive RSSL algorithms sometimes fails to select the optimal feature subsets?


\section{Conclusion and Future Work}

We presented a detailed quantitative analysis of the adaptive RSSL algorithm for an ensemble prediction problem. We support this analysis with deep theoretical (mathematical) explanation (formulation). The key important issues for the developed algorithm resides on four fundamental factors: generalization, flexibility, speed, and accuracy. We will explain each of these four factors.
We present a rigorous theoretical justification of our propose algorithm.
For now, we choose fixed number of attribute subset. However, the algorithm should evaluated based on the performance (accuracy) to determine the appropriate number (dimension) for single classifiers used in the ensemble learning. In addition, the adaptive RSSL algorithm is tested on a relatively small data sets. Our next step will be applying the developed algorithm on a big data sets.

Also, we show that the adaptive RSSL performs better than widely used ensemble algorithms even with the dependence of feature subsets.

Computational issues.
%
%


\bibliography{ICML-2015-EF-1}
\bibliographystyle{icml2015}

\end{document}